\documentclass{article}
\usepackage{spconf,amsmath,graphicx}
\usepackage{amsfonts}
\usepackage{booktabs}
\usepackage{subcaption}
\usepackage{lipsum}

\title{Cascading Unknown Detection with Known Classification for Open Set Recognition}
\name{Daniel Brignac, Abhijit Mahalanobis}
\address{University of Arizona\\
        Department of Electrical and Computer Engineering\\
        Tucson, AZ}
\begin{document}
%\ninept
%
\maketitle
\begin{abstract}
Deep learners tend to perform well when trained under the \emph{closed set} assumption but struggle when deployed under \emph{open set} conditions. This motivates the field of \emph{Open Set Recognition} in which we seek to give deep learners the ability to recognize whether a data sample belongs to the known classes trained on or comes from the surrounding infinite world. Existing open set recognition methods typically rely upon a single function for the dual task of distinguishing between knowns and unknowns as well as making known class distinction. This dual process leaves performance on the table as the function is not specialized for either task. In this work, we introduce \emph{Cascading Unknown Detection with Known Classification} (Cas-DC), where we instead learn specialized functions in a cascading fashion for both known/unknown detection and fine class classification amongst the world of knowns. Our experiments and analysis demonstrate that Cas-DC handily outperforms modern methods in open set recognition when compared using AUROC scores and correct classification rate at various true positive rates.
\end{abstract}


\section{Introduction}

Recent studies have demonstrated the capacity of deep learners to achieve or even surpass human-level performance, particularly in the image recognition domain. This performance is typically achieved under the \emph{closed set} assumption, however, in which the classes used for training the model are fixed and the model should only make predictions on this predefined set of classes. In practicality, the model may actually be deployed under \emph{open set} conditions where the classes used for training are only a subset of the infinite surrounding world and the model must be able to distinguish between these known, trained on classes and the encompassing open world.

\begin{figure}[t]
     \centering
     \includegraphics[width=0.80\linewidth]{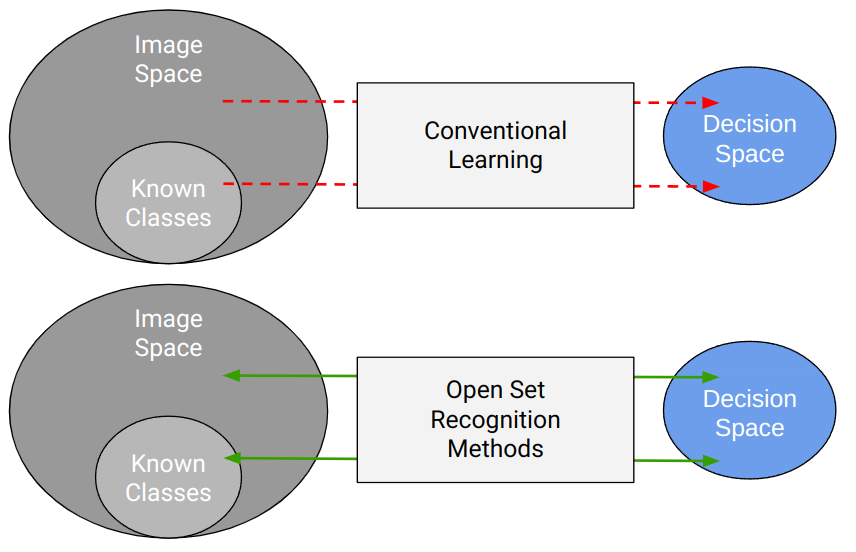}
     \caption{In conventional learning, classes from the known space (training classes) and the infinite surrounding image space map to the same decision space. Open set recognition methods allow the decision space to map back to the infinite surrounding image space by means of an "unknown" label.}
     \label{fig:Solution}
 \end{figure}

Conventionally, deep neural networks struggle under these open set conditions as they will confidently map unknown classes to the known class decision space \cite{DNNsFooled, OODBaseline} as demonstrated in Figure \ref{fig:Solution}. This motivates the study of \emph{Open Set Recognition} where we seek to discriminate between the world of \textit{knowns} the model is trained on and the surrounding infinite \textit{unknown} space. 

Open set recognition was first formalized in \cite{TowardOpenSet} and has since inspired an entire subfield of research. One of the first lines of work focused on an analysis of test time softmax scores \cite{OODBaseline} as classifiers trained under the closed set assumption tend to produce low softmax probabilities for samples belonging to the unknown space. \cite{OpenSetDeepNetworks} takes a similar route by extending the softmax layer to allow prediction of an unknown class. These softmax based methods still suffer in open set recognition due to the inherent limitations of training the networks under the closed set assumption \cite{DiscriminativeReciprocalPoints}.

Other methods take a generative approach \cite{Counterfactual, C2AE} in an attempt to generate samples belonging to the unknown world, or a distance-based approach \cite{NearestNeighbors, P-ODN} by thresholding a distance to the nearest known class. While these methods perform better than traditionally used softmax score analysis, they still do not perform to their maximum capability as they have no true representation for what the world of unknowns may resemble. 

Additionally, most current open set methods operate under the proposed setup of \cite{TowardOpenSet} in which a single function is given the task of distinguishing between knowns and unknowns and additionally making fine distinction amongst the world of knowns (i.e, classification). This leads to a function that may perform relatively well for this joint task, but is not specialized for either task leaving performance on the table.

To this end, we introduce our method \emph{Cascading Unknown Detection with Known Classification} (Cas-DC) to better address these shortcomings. In Cas-DC, we hypothesize that the known and unknown classes should clearly separate in the embedding space. This separation can be accomplished by training an embedding network with a representative set of the unknown world referred to as \emph{known unknowns} as in \cite{PMforOSR}. Each embedding space can then be represented by its respective prototype for best separation. Furthermore, we train a classifier network under the closed set assumption for discrimination amongst the world of knowns. At test time, we can determine if a sample belongs to the world of knowns or unknowns by setting a threshold on the distance to the unknown prototype, and if a sample is deemed as known, we can query the classifier to determine its class. This formulation of two specialized decision functions trained in a cascading fashion allows each to be an expert in their respective task leading to higher performance when combined together. 

% \section{Related Work}
% \lipsum[1-3]

\begin{figure*}[t]
     \centering
     \begin{subfigure}[b]{0.49\textwidth}
        \centering
        \includegraphics[width=0.9\linewidth]{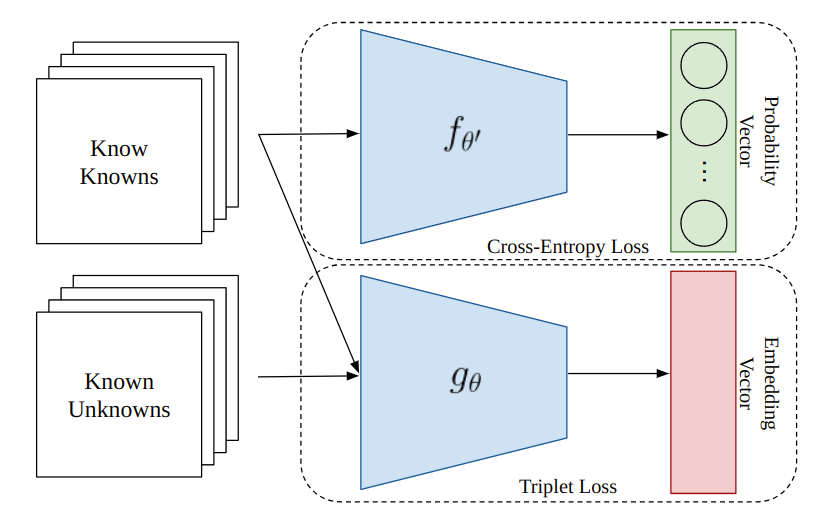}
        \caption{We train a classifier network $f_{\theta'}$ using only data consisting of known knowns with cross-entropy loss and an embedding network $g_\theta$ using known knowns data and a representative set of the unknown data termed known unknowns with triplet loss.}
        \label{fig:Training_Procedure}
     \end{subfigure}
     \hfill
     \begin{subfigure}[b]{0.49\textwidth}
    \centering
    \includegraphics[width=0.9\linewidth]{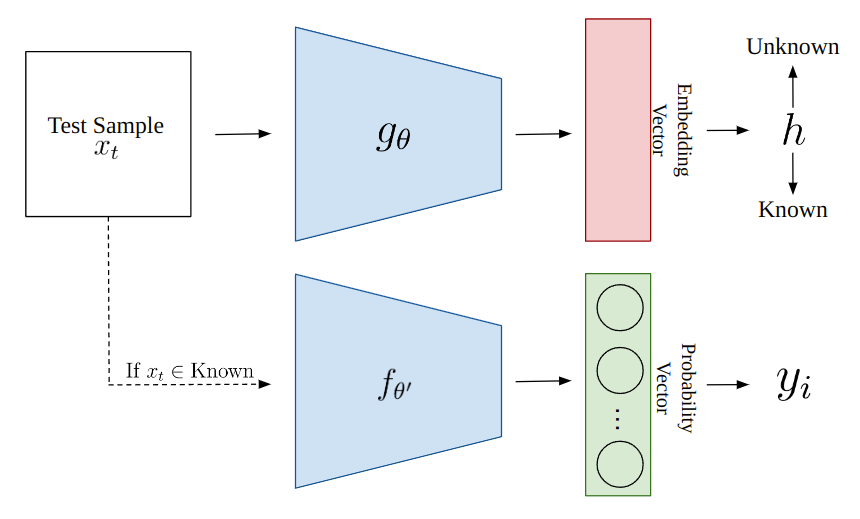}
    \caption{At test time we take a test sample and feed it to $g_\theta$ to get an embedding vector. We then feed that vector to the discriminator $h$ for known/unknown declaration. If known, we additionally feed the sample to the classifier $f_{\theta'}$ to obtain a fine class label.}
    \label{fig:Inference_Procedure}
     \end{subfigure}
     \hfill
        \caption{Cas-DC training procedure (a) and inference procedure (b).}
        \label{fig:Cas-DC_training_inference}
\end{figure*}

\section{Preliminaries}
We first establish the formalities of the open set recognition problem before formulating our proposed solution \cite{TowardOpenSet, OSRSurvey, DiscriminativeReciprocalPoints}. Suppose we are given a dataset $\mathcal{D}_{KK}$ of $n$ labeled data points we will refer to as \emph{known knowns}, namely $\mathcal{D}_{KK} = \{(x_1, y_1), ..., (x_n, y_n)\}$ where $y_i \in \{1, ..., C\}$ is the label for $x_i$ for $C$ unique class labels in $\mathcal{D}_{KK}$. At test time, we will perform inference on the larger test data $\mathcal{D}_T$ consisting of data from $\mathcal{D}_{KK}$ as well as data from an unknown set $\mathcal{D}_{UU}$, which we refer to as \emph{unknown unknowns}, whose labels $t_i \notin \{1, ..., C\}$. That is $\mathcal{D}_T = \mathcal{D}_{KK} \cup \mathcal{D}_{UU}$. We denote the embedding space of known category $k$ as $\mathcal{S}_k$ with corresponding open space $\mathcal{O}_k = \mathbb{R}^d - \mathcal{S}_k$ where $\mathbb{R}^d$ is the full embedding space consisting of known knowns and unknown unknowns. We further define the positive open space from other known knowns as $\mathcal{O}_k^{pos}$ and the remaining infinite space consisting of unknown unknowns as the negative open space $\mathcal{O}_k^{neg}$, that is $\mathcal{O}_k = \mathcal{O}_k^{pos} \cup \mathcal{O}_k^{neg}$.

We first introduce open set recognition for a single known class and then extend to the multi-class scenario. Given the data $\mathcal{D}_{KK}$, let samples from known category $k$ be positive training data occupying space $\mathcal{S}_k$, samples from other known classes be negative training data occupying space $\mathcal{O}_k^{pos}$, and all other samples from $\mathbb{R}^d$ be unknown data, $\mathcal{D}_{UU}$, occupying space $\mathcal{O}_k^{neg}$. Let $\psi_k: \mathbb{R}^d \rightarrow \{0, 1\}$ be a binary measurable prediction function which maps the embedding $x$ to label $y$ with the label for the class of interest $k$ being 1. In this 1-class scenario, we wish to optimize the discriminant binary function $\psi_k$ by minimizing the expected error $\mathcal{R}_k$ as 
\begin{equation}
    \underset{\psi_k}{\operatorname{arg min}} \{ \mathcal{R}_k = \mathcal{R}_o(\psi_k, \mathcal{O}_k^{neg}) + \alpha \mathcal{R}_\epsilon (\psi_k, \mathcal{S}_k \cup \mathcal{O}_k^{pos}) \}
\end{equation}
where $\mathcal{R}_o$ is the open space risk function, $\mathcal{R}_\epsilon$ is the empirical classification risk on the known data, and $\alpha$ is a regularization parameter.

We can extend to the multiclass recognition problem by incorporating multiple binary classification tasks and summing the expected risk category by category as
\begin{equation}
    \sum_{k=1}^C \mathcal{R}_o(\psi_k, \mathcal{O}_k^{neg})+ \alpha \sum_{k=1}^C \mathcal{R}_\epsilon (\psi_k, \mathcal{S}_k \cup \mathcal{O}_k^{pos})
\end{equation}
leading to the following formulation
\begin{equation} \label{eq: traditional_formulation}
    \underset{f \in \mathcal{H}}{\operatorname{arg min}} \{ \mathcal{R}_o(f, \mathcal{D}_{UU})  + \alpha  \mathcal{R}_\epsilon(f, \mathcal{D}_{KK})
    \}
\end{equation}
where $f: \mathbb{R}^d \rightarrow \mathbb{N}$ is a measurable multiclass recognition function. From this, we can see that solving the open set recognition problem is equivalent to minimizing the combination of the empirical classification risk on the labeled known data $\mathcal{D}_{KK}$ and open space risk on the unknown data $\mathcal{D}_{UU}$ simultaneously over the space of allowable recognition functions $\mathcal{H}$.

\section{Methodology}

\subsection{Cascading Unknown Detection with Known Classification}
In the traditional formulation of the open set recognition problem as described above, we assume a singular embedding space $\mathbb{R}^d$ consists of $N$ discriminant spaces for all known categories with all remaining space being the open space consisting of infinite unknowns. In formulating the framework of Cas-DC, we instead postulate that the embedding space $\mathbb{R}^d$ is composed of two disjoint spaces, namely a known space $\mathcal{S}_{known}$ and an unknown space $\mathcal{O}_{unknown}$. That is to say that all of $\mathcal{D}_{KK}$ belongs to the space $\mathcal{S}_{known}$ and all of $\mathcal{D}_{UU}$ belongs to the infinite surrounding open space $\mathcal{O}_{unknown}$. Thus, the open space is formulated as $\mathcal{O}_{unknown} = \mathbb{R}^d - \mathcal{S}_{known}$

Under this new assumption of the embedding space, we can now pose a new formulation of the open set recognition problem by introducing a cascading optimization procedure where we wish to optimize both a binary prediction function $h: \mathbb{R}^d \rightarrow \{0,1\}$ which maps the embedding of data $x$ to the label of known or unknown, and the classification function $f: x_i \rightarrow \mathbb{N}$ which maps the known data $x_i$ to their respective target label $y_i \in \{1, ..., N\}$ as
\begin{subequations} \label{eq: risk_equations}
\begin{align}
        \underset{h}{\operatorname{arg min}} & \text{ } \{\mathcal{R}_o (h, \mathbb{R}^d)\} \\
        \underset{f}{\operatorname{arg min}} & \text{ } \{\mathcal{R}_\epsilon(f, \mathcal{S}_{known})\}
\end{align}
\end{subequations}
where $\mathcal{R}_o$ is the open space risk and $\mathcal{R}_\epsilon$ is the empirical classification risk. Based on this formulation we can see that the first optimization procedure leads to another binary prediction function $h$ similar to the traditional formulation while the second procedure leads to a multiclass prediction function $f$.

All that remains now is to find a method that best creates the full embedding space $\mathbb{R}^d$ to give a simple discriminant function $h$ and obtain a high performing multiclass prediction function $f$.

\subsection{Embedding Separation of Knowns and Unknowns}
We first focus on the discrimination between knowns and unknowns in the embedding space $\mathbb{R}^d$. A deep neural network $g_\theta: x \rightarrow \mathbb{R}^d$ is used as an embedding network to obtain embedding vectors for all data $x \in \mathcal{D}_{KK} \cup \mathcal{D}_{UU}$. In order to enforce the separation between the spaces $\mathcal{S}_{known}$ and $\mathcal{O}_{unknown}$, the triplet loss \cite{FaceNet} is a natural choice of loss function to use when training $g_\theta$. One could consider using other contrastive learning methods such as contrastive loss \cite{ContrastiveLoss} or tuplet loss \cite{TupletLoss}, however, the choice to use triplet loss was made as contrastive loss only considers pairs and tuplet loss is a more general version of triplet loss.

With the triplet loss, we can treat all training data in $\mathcal{D}_{KK}$ as the positive samples. For negative samples, we now need to find a representation of $\mathcal{D}_{UU}$ for modeling the space $\mathcal{O}_{unknown}$. Of course this open space and therefore this dataset is infinite, but we can use a representative set of $\mathcal{D}_{UU}$ we refer to as \emph{known unknowns}, $\mathcal{D}_{KU} \subseteq \mathcal{D}_{UU}$, to train $g_\theta$ for embedding space separation of knowns and unknowns. The choice to use a representative training set $\mathcal{D}_{KU}$ to represent the entire world of unknowns is taken from out-of-distribution detection literature \cite{ODIN, Outlier-Exposure, OODUnifiedFramework, StatisticalFrameworkOOD}. Similar to \cite{Outlier-Exposure}, we only use $\mathcal{D}_{KU}$ during the training stage. At inference, we evaluate on the test data $\mathcal{D}_T = \mathcal{D}_{KK} \cup \mathcal{D}_{UU}$.

Now armed with the known training set $\mathcal{D}_{KK}$ and representative unknown training set $\mathcal{D}_{KU}$, we can formalize use of the triplet loss to train $g_\theta$ as 
\begin{equation} \label{eq: triplet_loss}
    \mathcal{L}_{g_\theta} = \sum_{i=1}^n || g_\theta(x_i^a) - g_\theta(x_i^{KK}) ||_2^2 - || g_\theta(x_i^a) - g_\theta(x_i^{KU}) ||_2^2 + \beta
\end{equation}
where $x_i^a$ is a known known anchor, $x_i^{KK}$ is a known known positive sample, $x_i^{KU}$ is a known unknown negative sample, and $\beta$ is a margin that is enforced between the positive and negative pairs.

\subsection{Discrimination Between Knowns and Unknowns}
With a binary discriminant embedding space $\mathbb{R}^d$ now at hand, we must now develop the discriminant function $h$ to differentiate between knowns and unknowns. As such, we draw inspiration from \cite{DistanceBasedImageClassification, IncrementalLearningNCM, OpenSetDeepNetworks} by measuring the distance to the embedding prototypes for known/unknown discrimination. We represent each of the known and unknown clusters in the embedding space by their respective prototype determined by taking the means of the known knowns, $\mu_{KK}$, and known unknowns, $\mu_{KU}$, in the embedding space.

We then measure the Euclidean distance to $\mu_{KU}$ and set a threshold for final determination of whether a test sample is known or unknown. Thus, the binary function $h$ takes the form
\begin{equation}
h =
   \begin{cases} 
      known  & \text{if } d(g_\theta(x_t), \mu_{KU}) > \tau\\
      unknown & \text{if } d(g_\theta(x_t), \mu_{KU}) \leq \tau\\
   \end{cases}
\end{equation}
where $x_t$ is a test sample from $\mathcal{D}_T$, $d(g_\theta(x_t), \mu_{KU}) = || g_\theta(x_t) - \mu_{KU} ||_2^2$ is the Euclidean distance between the embedding of $x_t$ and the known unknown prototype $\mu_{KU}$ and $\tau$ is a threshold.

\subsection{Management of Open Space Risk}

In theory, the open space $\mathcal{O}_{unknown}$ is infinite making for difficult management of the open space risk $\mathcal{R}_o$. We instead opt to indirectly bound this open space for easier management of $\mathcal{R}_o$ as a direct bounding would be nearly impossible due to the infinite nature of $\mathcal{O}_{unknown}$. By enforcing the distance between samples from $\mathcal{S}_{known}$ and $\mathcal{O}_{unknown}$ to be outside some predefined margin of separation we are able to indirectly bound $\mathcal{O}_{unknown}$. This bounding procedure gives rise to Eq. \ref{eq: triplet_loss}  which enforces the distance between samples from the known knowns and known unknowns to be greater than or equal to the margin $\beta$.

The use of $\mathcal{D}_{KK}$ and $\mathcal{D}_{KU}$ in the training of $g_\theta$ for embedding space separation gives rise to the bounding spaces $\mathcal{B}_{known}$ and $\mathcal{B}_{unknown}$ respectively. Ideally, these spaces would be completely separable in $\mathbb{R}^d$, but in practicality there will be some overlap in the margin region. By representing each bounding space by its prototype as described above, we are able to achieve greater separation in $\mathbb{R}^d$. As a result, training with triplet loss for separation between $\mathcal{B}_{known}$ and $\mathcal{B}_{unknown}$ and further representing each bounding region with its appropriate prototype for final binary prediction can be viewed as managing the open space risk $\mathcal{R}_o(h, \mathbb{R}^d)$ in Eq. \ref{eq: risk_equations}.

\subsection{Distinction Amongst Knowns}
The last remaining step is now developing a way to best identify which known class a sample belongs to for reduction of the empirical classification risk $\mathcal{R}_\epsilon$. In order to distinguish fine class labels amongst the world of knowns, we train a separate deep neural network $f_{\theta'}$ using cross-entropy loss in parallel with the embedding network $g_\theta$. As $f_{\theta'}$ is only concerned with classification of the knowns, we only use the data from $\mathcal{D}_{KK}$ to train the classifier. Figure \ref{fig:Training_Procedure} shows the full training procedure for training the multiclass prediction function $f_{\theta'}$ and the embedding network $g_\theta$.

At the inference stage, we only query $f_{\theta'}$ for a fine class label if the binary discriminant function $h$ predicts that a test sample $x_t$ belongs to the known space $\mathcal{S}_{known}$. Otherwise, $x_t$ is assigned to the world of unknowns. Figure \ref{fig:Inference_Procedure} gives an overview for the entire inference stage.

\begin{table*}[t]
    %\small
    \centering
    \scriptsize
    \begin{tabular}{c|c|c|c|c|c|c}
    \toprule
        \textbf{Method} & \textbf{MNIST} & \textbf{SVHN} & \textbf{CIFAR10} & \textbf{CIFAR+10} & \textbf{CIFAR+50} & \textbf{Tiny-Imagenet}\\
        \midrule
        \textbf{Counter-Factual Images} & $0.9857 \pm 0.006$ & $0.9081 \pm 0.008$ & $0.6999 \pm 0.006$ & $0.8251 \pm 0.004$ & $0.8168 \pm 0.001$ & $0.5734 \pm 0.007$ \\
        \textbf{Outlier Exposure*} & $0.9814 \pm 0.002$ & $0.9056 \pm 0.011$ & $0.9354 \pm 0.014$ & $0.8482 \pm 0.018$ & $\mathbf{0.9570 \pm 0.002}$ & $0.7215 \pm 0.006$ \\
        \textbf{Class Anchor Clustering} & $0.8187 \pm 0.011$ & $0.9038 \pm 0.015$ & $0.7156 \pm 0.002$ & $0.7425 \pm 0.013$ & $0.7721 \pm 0.002$ & $0.5452 \pm 0.036$ \\
        \textbf{Good Classifier} & $0.9894 \pm 0.001$ & $0.9058 \pm 0.012$ & $0.7479 \pm 0.008$ & $0.7734 \pm 0.014$ & $0.7720 \pm 0.002$ & $0.6291 \pm 0.016$ \\
        \textbf{ARPL+CS} & $0.9900 \pm 0.001$ & $0.9342 \pm 0.005$ & $0.7813 \pm 0.002$ & $0.8346 \pm 0.005$ & $0.8241 \pm 0.004$ & $0.6402 \pm 0.023$\\
        
        \midrule
        \textbf{Cas-DC (Ours)*} & $\mathbf{0.9930 \pm 0.001}$ & $\mathbf{0.9498 \pm 0.017}$ & $\mathbf{0.9752 \pm 0.002}$ & $\mathbf{0.9264 \pm 0.008}$ & $0.9475 \pm 0.002$ & $\mathbf{0.7240\pm 0.020}$ \\
    \bottomrule
    \end{tabular}
    \caption{Reported AUROC score means and standard deviations for each tested method for the various tested datasets averaged over 3 runs. The character (*) signifies the use of known unknowns during the training procedure.}
    \label{tab:AUROC}
\end{table*}

\begin{table*}[t]
    %\small
    \centering
    \scriptsize
    \begin{tabular}{c|c|c|c|c|c|c}
    \toprule
        \textbf{Method} & \textbf{MNIST} & \textbf{SVHN} & \textbf{CIFAR10} & \textbf{CIFAR+10} & \textbf{CIFAR+50} & \textbf{Tiny-Imagenet}\\
        \midrule
        %\textbf{Counter-Factual Images*} & $0.0 \pm 0.0$ & $0.0 \pm 0.0$ & $0.7363 \pm 0.003$ & $0.8946 \pm 0.001$ & $0.89475 \pm 0.001$ & $0.3873 \pm 0.014$ \\
        \textbf{Class Anchor Clustering} & $0.9950 \pm 0.0003$ & $0.8751 \pm 0.056$ & $0.688 \pm 0.009$ & $\mathbf{0.8869 \pm 0.004}$ & $0.8805 \pm 0.007$ & $0.3773 \pm 0.038$ \\
        \textbf{Outlier Exposure*} & $0.9958 \pm 0.001$ & $0.9549 \pm 0.004$ & $0.7439 \pm 0.024$ & $0.8525 \pm 0.021$ & $\mathbf{0.9022 \pm 0.0006}$ & $0.6183 \pm 0.021$ \\
        \textbf{Good Classifier} & $0.9599 \pm 0.005$ & $0.7260 \pm 0.021$ & $0.5650 \pm 0.001$ & $0.5731 \pm 0.012$ & $0.5694 \pm 0.003$ & $0.5263 \pm 0.002$ \\
        \textbf{ARPL+CS} & $0.9508 \pm 0.013$ & $0.8340 \pm 0.002$ & $0.6571 \pm 0.002$ & $0.8233 \pm 0.002$ & $0.5821 \pm 0.004$ & $0.1732 \pm 0.004$\\
        \midrule
        \textbf{Cas-DC (Ours)*} & $\mathbf{0.9961 \pm 0.001}$ & $\mathbf{0.9550 \pm 0.008}$ & $\mathbf{0.6947 \pm 0.0001}$ & $0.8670 \pm 0.004$ & $0.8700 \pm 0.0008$ & $\mathbf{0.6194 \pm 0.029}$\\
    \bottomrule
    \end{tabular}
    \caption{Reported CCR at 95\% TPR score means and standard deviations for each tested method for the various tested datasets averaged over 3 runs. The character (*) signifies the use of known unknowns during the training procedure.}
    \label{tab:CCR}
\end{table*}

\section{Experiments and Results}

\subsection{Experimental Setup}

\textbf{Datasets.} We test on six commonly used datasets in open set recognition literature. In the MNIST and SVHN datasets, we randomly choose 6 classes as known and the remaining 4 classes as unknown. Each of the CIFAR datasets is taken from either CIFAR10 or a combination of CIFAR10 and CIFAR100. For CIFAR10 experiments, all experiments are performed by treating the 6 non-vehicle classes as known classes and the remaining 4 vehicle classes as the unknown (i.e., open) classes. CIFAR+M experiments takes the 4 vehicle classes from CIFAR10 as known and randomly samples from M disjoint classes (i.e., non-vehicle classes) from the CIFAR100 dataset. Lastly, in Tiny-Imagenet experiments we randomly choose 20 classes as the known classes and treat all other 180 classes as unknown.

\textbf{Metrics.} We use the standard area under the ROC curve (AUROC) as the main metric when evaluating the performance of all compared methods. A draw back of AUROC as commonly reported in open set trials, is it only takes into consideration known/unknown discrimination. A good open set recognizer should be able to additionally discriminate amongst the knowns given that a sample is predicted to be known. For this reason we additionally report the correct classification rate (CCR) at 95\% true positive rate (TPR) of known detection.

\textbf{Compared Methods.} We compare our method, Cas-DC, to five open set recognition methods that are most comparable in regards to methodology. Counter-factual images \cite{Counterfactual} uses a GAN to generate counter examples to the known class which are then treated as the unknown class and used to train a "$K+1$" classifier where the $(K+1)^{th}$ class is the unknown class. Similar to our method, outlier exposure \cite{Outlier-Exposure} uses known unknowns to fine-tine a classifier for open set detection based on the output softmax score. Class anchor clustering (CAC) \cite{CAC} poses a new loss function to entice each of the distinct known classes to cluster around their respective standard basis vector so that the unknown classes will then occupy the remaining open space. A distance threshold is then used for distinct known or unknown discrimination similar to Cas-DC. Adversarial Reciprocal Point Learning + confusion samples (ARPL+CS) \cite{ARPL+CS} learns reciprocal points for each known class open space while simultaneously using a generator to generate confusing training samples to encourage known class separation in the latent space and uses a distance measure to the furthest reciprocal point to obtain a probability of belonging to a particular known class. Lastly, \cite{GoodClassifier} propose that the best open set recognition model is simply one that is a Good Classifier for the closed-set scenario. With this good closed-set classifier at hand, an analysis of the maximum logit score produced by a sample is used in the final determination of distinct known or unknown.

\textbf{Setup.} For all methods, we train on the dataset splits described above. For neural network architectures, we use Resnet18 in all tested methods for fairest comparisons except in counterfactual images and CAC. We keep the architectures unchanged in both of these methods as the former used a specific generator and discriminator for best GAN performance and the latter did not allow simplistic modulation with a Resnet encoder. Besides described architecture changes, all other hyperparemeters for compared methods remain unchanged. All methods are trained via SGD with standard L2 regularization. For Cas-DC, the margin of separation $\beta$ in Eq. \ref{eq: triplet_loss} is found empirically and a combination of semihard and hard negative mining are used for finding triplets. Results for best $\beta$ selection are found in the supplementary material. Lastly, we use half of unknown classes for all datasets as the training set $\mathcal{D}_{KU}$ in Cas-DC. We additionally report various percentages of full unknown dataset used for $\mathcal{D}_{KU}$ in the supplementary material.

\subsection{Results Comparison}

We first evaluate the performance of Cas-DC vs. all other compared methods from an AUROC standpoint. Table \ref{tab:AUROC} shows AUROC results averaged across 3 runs for all methods. We observe that Cas-DC either outperforms or is very competitive with all compared methods for all datasets. This can be attributed to Cas-DC's specialized function $h$ for declaration of knowns and unknowns whereas all other methods use a singular function for both known/unknown discrimination and known class distinction as is commonly done in the traditional formulation of the open set recognition problem in Eq. \ref{eq: traditional_formulation}.

\begin{figure}
    \centering
    \includegraphics[width=0.80\linewidth]{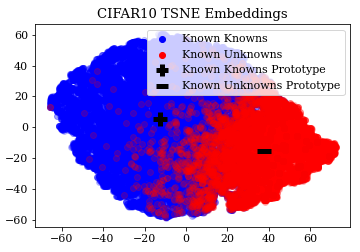}
    \caption{CIFAR10 TSNE plot of the embedding space.}
    \label{fig:TSNE Plot}
\end{figure}

Additionally, Cas-DC's $h$ discriminator is further assisted by clear known and unknown separation in the embedding space $\mathbb{R}^d$ as initially hypothesized by means of the triplet loss. We can confirm this by analyzing the TSNE plot of the embeddings produced by $g_\theta$ as done in Figure \ref{fig:TSNE Plot} for the CIFAR10 data split. Of course, we observe an overlap region where discrimination between knowns and unknowns can prove challenging, but by representing each embedding cluster by its respective prototype, we are able to achieve better separation leading to a more favorable AUROC performance.

We now evaluate the performance of Cas-DC against all other compared methods from a CCR standpoint. Table \ref{tab:CCR} reports the CCR at 95\% TPR for all methods except Counter-Factual Images. We do not report results for Counter-Factual images due to the inherent nature of using a "$K+1$" classifier (i.e., the "$K+1$" classifier is not dependent on known/unknown discrimination as course distinction is based on discriminator scores and fine distinction amongst the "$K+1$" classes is based on separate classifier scores). We overall observe that Cas-DC is mostly competitive with all other tested methods, but in particular performs exceptionally well on Tiny-Imagenet along with outlier exposure. The clear superiority of Cas-DC and outlier exposure on Tiny-Imagenet can be attributed to the use of known unknowns during the training phase as known unknowns allow us to learn a representative space for all unknowns.

\begin{figure}
    \centering
    \includegraphics[width=0.80\linewidth]{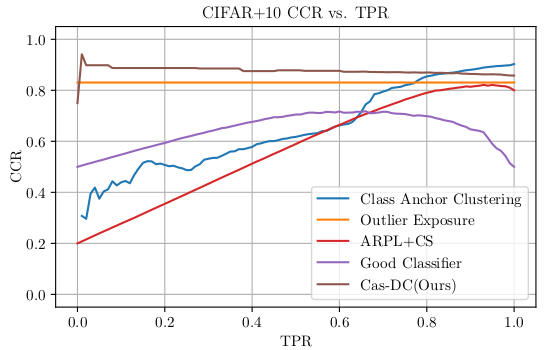}
    \caption{CIFAR+10 CCR for varying TPR.}
    \label{fig:CCR_vs_TPR_curves}
\end{figure}

While Cas-DC remains competitive in all other datasets in regards to CCR at 95\% TPR, we question if this is true for all operating TPRs. To answer this, we plot the CCR against various TPRs for the CIFAR+10 dataset in Figure \ref{fig:CCR_vs_TPR_curves}. From this, we make multiple interesting observations. Firstly, we can observe that Cas-DC is, in general, more stable than most of the compared methods. Again, this can be attributed to having a specialized classifier capable of consistent performance regardless of the number of known declarations. Secondly, Figure \ref{fig:CCR_vs_TPR_curves} suggests that at nearly all other operating TPRs, Cas-DC is in fact superior. This would suggest that Cas-DC is the superior method in scenarios where higher TPRs can be waived. We make note of outlier exposure having a steady correct classification rate regardless of the TPR used. This is attributed to outlier exposure's independent nature of assigning an unknown label vs. a known label which can be advantageous in scenarios where a lower, but still strong CCR is desired.

Lastly, we analyze the impact of the margin of separation $\beta$ and number of classes used for known unknowns during training in Table \ref{tab:varying_beta} and Table \ref{tab:varying_ku} respectively. We observe that generally as we increase the margin of separation, we slowly see a degradation in performance. An increase in the margin corresponds to an increase in the distance between a positive sample and a negative sample in the triplet loss. Thus, when samples become harder to distinguish between positive and negative, we incur more risk with a higher margin leading to worse performance.

In regards to the number of samples used for known unknowns during training, naturally we expect performance to increase as we use more unknowns during training. We can clearly see this behavior in Table \ref{tab:varying_ku}, but using more unknowns for training might be unfeasible in certain scenarios. Thus, we look where performance begins to saturate and arrive as the conclusion that half of the unknowns suffices for our experimentation.

\begin{table}[t]
    %\small
    \centering
    \scriptsize
    \begin{tabular}{c|c|c|c}
    \toprule
        \textbf{Value} & \textbf{SVHN} & \textbf{CIFAR+50} & \textbf{Tiny-Imagenet}\\
        \midrule
        $\beta=0.1$ & $0.9498 \pm 0.017$ & $0.9475 \pm 0.002$ & $0.7240\pm 0.020$ \\
        $\beta=0.2$ & $0.8759 \pm 0.015$ & $0.9112 \pm 0.016$ & $0.6048 \pm 0.004$ \\
        $\beta=0.5$ & $0.8476 \pm 0.020$ & $0.8977 \pm 0.010$ & $0.5945 \pm 0.009$  \\
        $\beta=0.9$ & $0.5185 \pm 0.006$ & $0.8855 \pm 0.010$ & $0.5763 \pm 0.004$ \\
        $\beta=1.0$ & $0.5092 \pm 0.002$ & $0.8844 \pm 0.001$ & $0.5601 \pm 0.015$ \\
    \bottomrule
    \end{tabular}
    \caption{Effect of various $\beta$ values with respect to AUROC performance averaged across 3 runs. Above results use a half of the unknowns as known unknowns.}
    \label{tab:varying_beta}
\end{table}

\begin{table}[t]
    %\small
    \centering
    \scriptsize
    \begin{tabular}{c|c|c|c}
    \toprule
        \textbf{Unknown \%} & \textbf{SVHN} & \textbf{CIFAR+50} & \textbf{Tiny-Imagenet}\\
        \midrule
        $10\%$ & $0.8299 \pm 0.007$ & $0.8733 \pm 0.007$ & $0.5245 \pm 0.002$ \\
        $20\%$ & $0.8239 \pm 0.004$ & $0.9161 \pm 0.003$ & $0.6063 \pm 0.019$ \\
        $50\%$ & $0.9498 \pm 0.017$ & $0.9475 \pm 0.002$ & $0.7240\pm 0.020$  \\
        $90\%$ & $0.9630 \pm 0.010$ & $0.9519 \pm 0.002$ & $0.7301 \pm 0.020$ \\
    \bottomrule
    \end{tabular}
    \caption{Effect of various percentages of total unknowns used as known unknowns during training in regards to AUROC performance averaged across 3 runs. Above results use the best configuration for margin of separation $\beta$.}
    \label{tab:varying_ku}
\end{table}

\section{Conclusion}
In this work, we introduce our method Cas-DC for open set recognition. Cas-DC benefits from having two specialized functions for known and unknown discrimination as well as fine class distinction amongst knowns. This allows each function to be an expert for their respective task allowing for top tier performance compared to that of traditional open set recognition methods where a single function is used for both known/unknown discrimination and fine class distinction. Additionally, by using a representative set of the unknowns termed \emph{known unknowns}, we are able to train an embedding network for distinct separation between knowns and unknowns in the embedding space allowing for easy discrimination. Our experiments show that we outperform modern open set recognition methods in not only known/unknown discrimination, but also correct classification amongst the knowns. We summarize our conclusions as follows:
\begin{itemize}
    \item We present Cas-DC, a novel method for open set recognition that decouples the binary discriminator from known class distinction using a metric learning approach.
    \item We show that using a representative set of the unknowns, termed \emph{known unknowns}, during training yields superior performance for open set recognition.
    \item We demonstrate that having a specialized function for each of known/unknown distinction and known class discrimination yields overall superior performance.
\end{itemize}

\bibliographystyle{IEEEbib}
\bibliography{main}

\begin{thebibliography}{10}

\bibitem{DNNsFooled}
Anh Nguyen, Jason Yosinski, and Jeff Clune,
\newblock ``Deep neural networks are easily fooled: High confidence predictions for unrecognizable images,''
\newblock in {\em Proceedings of the IEEE conference on computer vision and pattern recognition}, 2015, pp. 427--436.

\bibitem{OODBaseline}
Dan Hendrycks and Kevin Gimpel,
\newblock ``A baseline for detecting misclassified and out-of-distribution examples in neural networks,''
\newblock {\em Proceedings of International Conference on Learning Representations}, 2017.

\bibitem{TowardOpenSet}
Walter~J. Scheirer, Anderson de~Rezende~Rocha, Archana Sapkota, and Terrance~E. Boult,
\newblock ``Toward open set recognition,''
\newblock {\em IEEE Transactions on Pattern Analysis and Machine Intelligence}, vol. 35, no. 7, pp. 1757--1772, 2013.

\bibitem{OpenSetDeepNetworks}
Abhijit Bendale and Terrance~E Boult,
\newblock ``Towards open set deep networks,''
\newblock in {\em Proceedings of the IEEE conference on computer vision and pattern recognition}, 2016, pp. 1563--1572.

\bibitem{DiscriminativeReciprocalPoints}
Guangyao Chen, Limeng Qiao, Yemin Shi, Peixi Peng, Jia Li, Tiejun Huang, Shiliang Pu, and Yonghong Tian,
\newblock ``Learning open set network with discriminative reciprocal points,''
\newblock in {\em Computer Vision--ECCV 2020: 16th European Conference, Glasgow, UK, August 23--28, 2020, Proceedings, Part III 16}. Springer, 2020, pp. 507--522.

\bibitem{Counterfactual}
Lawrence Neal, Matthew Olson, Xiaoli Fern, Weng-Keen Wong, and Fuxin Li,
\newblock ``Open set learning with counterfactual images,''
\newblock in {\em Proceedings of the European Conference on Computer Vision (ECCV)}, 2018, pp. 613--628.

\bibitem{C2AE}
Poojan Oza and Vishal~M Patel,
\newblock ``C2ae: Class conditioned auto-encoder for open-set recognition,''
\newblock in {\em Proceedings of the IEEE/CVF Conference on Computer Vision and Pattern Recognition}, 2019, pp. 2307--2316.

\bibitem{NearestNeighbors}
Pedro~R Mendes~J{\'u}nior, Roberto~M De~Souza, Rafael de~O Werneck, Bernardo~V Stein, Daniel~V Pazinato, Waldir~R de~Almeida, Ot{\'a}vio~AB Penatti, Ricardo da~S Torres, and Anderson Rocha,
\newblock ``Nearest neighbors distance ratio open-set classifier,''
\newblock {\em Machine Learning}, vol. 106, no. 3, pp. 359--386, 2017.

\bibitem{P-ODN}
Yu~Shu, Yemin Shi, Yaowei Wang, Tiejun Huang, and Yonghong Tian,
\newblock ``P-odn: Prototype-based open deep network for open set recognition,''
\newblock {\em Scientific reports}, vol. 10, no. 1, pp. 7146, 2020.

\bibitem{PMforOSR}
Walter~J. Scheirer, Lalit~P. Jain, and Terrance~E. Boult,
\newblock ``Probability models for open set recognition,''
\newblock {\em IEEE Transactions on Pattern Analysis and Machine Intelligence}, vol. 36, no. 11, pp. 2317--2324, 2014.

\bibitem{OSRSurvey}
Chuanxing Geng, Sheng-jun Huang, and Songcan Chen,
\newblock ``Recent advances in open set recognition: A survey,''
\newblock {\em IEEE transactions on pattern analysis and machine intelligence}, vol. 43, no. 10, pp. 3614--3631, 2020.

\bibitem{FaceNet}
Florian Schroff, Dmitry Kalenichenko, and James Philbin,
\newblock ``Facenet: A unified embedding for face recognition and clustering,''
\newblock in {\em Proceedings of the IEEE conference on computer vision and pattern recognition}, 2015, pp. 815--823.

\bibitem{ContrastiveLoss}
Prannay Khosla, Piotr Teterwak, Chen Wang, Aaron Sarna, Yonglong Tian, Phillip Isola, Aaron Maschinot, Ce~Liu, and Dilip Krishnan,
\newblock ``Supervised contrastive learning,''
\newblock {\em Advances in neural information processing systems}, vol. 33, pp. 18661--18673, 2020.

\bibitem{TupletLoss}
Kihyuk Sohn,
\newblock ``Improved deep metric learning with multi-class n-pair loss objective,''
\newblock {\em Advances in neural information processing systems}, vol. 29, 2016.

\bibitem{ODIN}
Shiyu Liang, Yixuan Li, and Rayadurgam Srikant,
\newblock ``Enhancing the reliability of out-of-distribution image detection in neural networks,''
\newblock {\em arXiv preprint arXiv:1706.02690}, 2017.

\bibitem{Outlier-Exposure}
Dan Hendrycks, Mantas Mazeika, and Thomas Dietterich,
\newblock ``Deep anomaly detection with outlier exposure,''
\newblock {\em arXiv preprint arXiv:1812.04606}, 2018.

\bibitem{OODUnifiedFramework}
Kimin Lee, Kibok Lee, Honglak Lee, and Jinwoo Shin,
\newblock ``A simple unified framework for detecting out-of-distribution samples and adversarial attacks,''
\newblock {\em Advances in neural information processing systems}, vol. 31, 2018.

\bibitem{StatisticalFrameworkOOD}
Matan Haroush, Tzviel Frostig, Ruth Heller, and Daniel Soudry,
\newblock ``A statistical framework for efficient out of distribution detection in deep neural networks,''
\newblock in {\em International Conference on Learning Representations}, 2022.

\bibitem{DistanceBasedImageClassification}
Thomas Mensink, Jakob Verbeek, Florent Perronnin, and Gabriela Csurka,
\newblock ``Distance-based image classification: Generalizing to new classes at near-zero cost,''
\newblock {\em IEEE transactions on pattern analysis and machine intelligence}, vol. 35, no. 11, pp. 2624--2637, 2013.

\bibitem{IncrementalLearningNCM}
Marko Ristin, Matthieu Guillaumin, Juergen Gall, and Luc Van~Gool,
\newblock ``Incremental learning of ncm forests for large-scale image classification,''
\newblock in {\em Proceedings of the IEEE conference on computer vision and pattern recognition}, 2014, pp. 3654--3661.

\bibitem{CAC}
Dimity Miller, Niko Sunderhauf, Michael Milford, and Feras Dayoub,
\newblock ``Class anchor clustering: A loss for distance-based open set recognition,''
\newblock in {\em Proceedings of the IEEE/CVF Winter Conference on Applications of Computer Vision}, 2021, pp. 3570--3578.

\bibitem{ARPL+CS}
Guangyao Chen, Peixi Peng, Xiangqian Wang, and Yonghong Tian,
\newblock ``Adversarial reciprocal points learning for open set recognition,''
\newblock {\em IEEE Transactions on Pattern Analysis and Machine Intelligence}, vol. 44, no. 11, pp. 8065--8081, 2021.

\bibitem{GoodClassifier}
Sagar Vaze, Kai Han, Andrea Vedaldi, and Andrew Zisserman,
\newblock ``Open-set recognition: A good closed-set classifier is all you need,''
\newblock in {\em International Conference on Learning Representations}, 2022.

\end{thebibliography}

\end{document}